\documentclass[11pt]{article}

\usepackage[final]{acl}

\usepackage{times}
\usepackage{latexsym}
\usepackage[T1]{fontenc}

\usepackage[utf8]{inputenc}
\usepackage{microtype}

\usepackage{inconsolata}

\usepackage{graphicx}
\usepackage{booktabs}
\usepackage{multirow}
\usepackage{subfigure}
\usepackage{subcaption}
\usepackage{bm}
\usepackage{paralist}
\usepackage{annotates}

\title{Comparing Architectures for Supervised Political Scaling}

\author{Anna Golub \and Sebastian Padó \\
    Institute for Natural Language Processing (IMS) \\
    University of Stuttgart, Germany \\
  \texttt{\{anna.golub,sebastian.pado\}@ims.uni-stuttgart.de}}

\begin{document}
\maketitle
\begin{abstract}
Text scaling, the task of positioning political actors on an ideological axis, is fundamental to political science. To ease the need for manual analysis, various NLP methods have been proposed for this task, including classification- and regression-based approaches, showing successes as well as limitations. The goal of our paper is to consolidate the state of the art in this area. We ask two questions: (a) Can the performance of scaling methods be improved by predicting scales not individually but jointly? (b) Is there a middle ground between classification and regression? 
\end{abstract}

\section{Introduction}


\textit{Text scaling}, the task of extracting the stances of political actors from written documents or speeches and mapping them to scores on an ideological axis, is fundamental to political text analysis \citep{laver2003-wordscores,Diermeier_2012,Barbera_2015}.
Representing party positions in this way allows quantitatively measuring the differences between them, which is instrumental in understanding voters' behavior during elections as well as the parties' strategies once in office \citep{benoit_laver_ppmd_2006}. 

Determining a gold standard for party positions is challenging. A well-known source, the Chapel Hill Survey \citep{ROVNY2025102981} is based on directly querying experts. Alternatively, the Manifesto Research on Political Representation (MARPOR) project\footnote{Previously known as the Comparative Manifesto Project (CMP), \url{https://manifestoproject.wzb.eu/}} grounds party positioning directly in texts. They have collected over 3200 manifestos (regarded as the most comprehensive source on party policy) and annotated the statements in them according to a fine-grained ontology. The overall party positions can then be obtained by aggregating category frequencies (see \S\ref{sec:methods-label-aggr} for details). The most well-known scale arising from this work is \textbf{RILE}, or the Standard Right-Left Scale, reflecting mainly positions on economic policy \citep{volkens-rile-2013, rile_budge_2013}.

To alleviate the cost of manual annotation, the scores can be estimated automatically with the help of NLP methods. Early word frequency-based statistical methods \citep{laver2003-wordscores, slapin2008-wordfish} already revealed the potential of computerized approaches for estimating party positions. Later, those results were extended and improved upon by methods from distributional semantics \citep{glavas-etal-2017, Rheault_Cochrane_party2vec}, Transformer-based sentence embeddings \cite{ceron-2022-dis-similarity,nikolaev-etal-2023} and large language models (LLMs) \cite{benoit2025llms}. 
 

While encouraging, the results of these studies unveil that the general approaches to political scaling suffer from two interrelated problems. Firstly, despite the broad consensus in political science that multiple axes are necessary to adequately capture party positions \citep{koedam_binding_steenbergen_2025}, most computational work focuses on positioning on a \textit{single scale} (usually RILE). As a result, models arguably need to learn to ignore a lot of information in the input instead of using it to their advantage. Secondly, political documents, such as election manifestos, tend to be \textit{too long} for current Transformer models to process in one and need to be subdivided in some manner.

To address these shortcomings, we systematically compare the performance of different approaches, focusing on two research questions:
\begin{description}
\item[RQ1: Does joint prediction improve scaling?] The \textbf{GAL-TAN}, ranging from the Green-Alternative-Liberal to the Traditional-Authoritarian-Nationalist extreme \citep{MarksSteenbergen2004}, 
captures socio-cultural views that complement RILE's economic perspective. Still, to our knowledge, there is no work on predicting GAL-TAN positions. What is more, despite being orthogonal in theory, the empirical RILE and GAL-TAN scores are often found to be interdependent. Left-wing parties gravitate toward \textit{GAL} social policies, and right-wing views tend to co-occur \textit{TAN} ones, whereas the remaining two combinations are much less frequent \citep{jahn2011,Brigevich2017-GALTAN, wagner2021party}. In fact, in the MARPOR dataset we use for training (see Section \ref{sec:data} for details), the correlation between RILE and GAL-TAN scores is $\rho=0.48$. On this basis, we expect that predicting RILE and GAL-TAN together with a joint model will improve performance.

\begin{figure}[tb!]
    \centering
    \includegraphics[width=\linewidth]{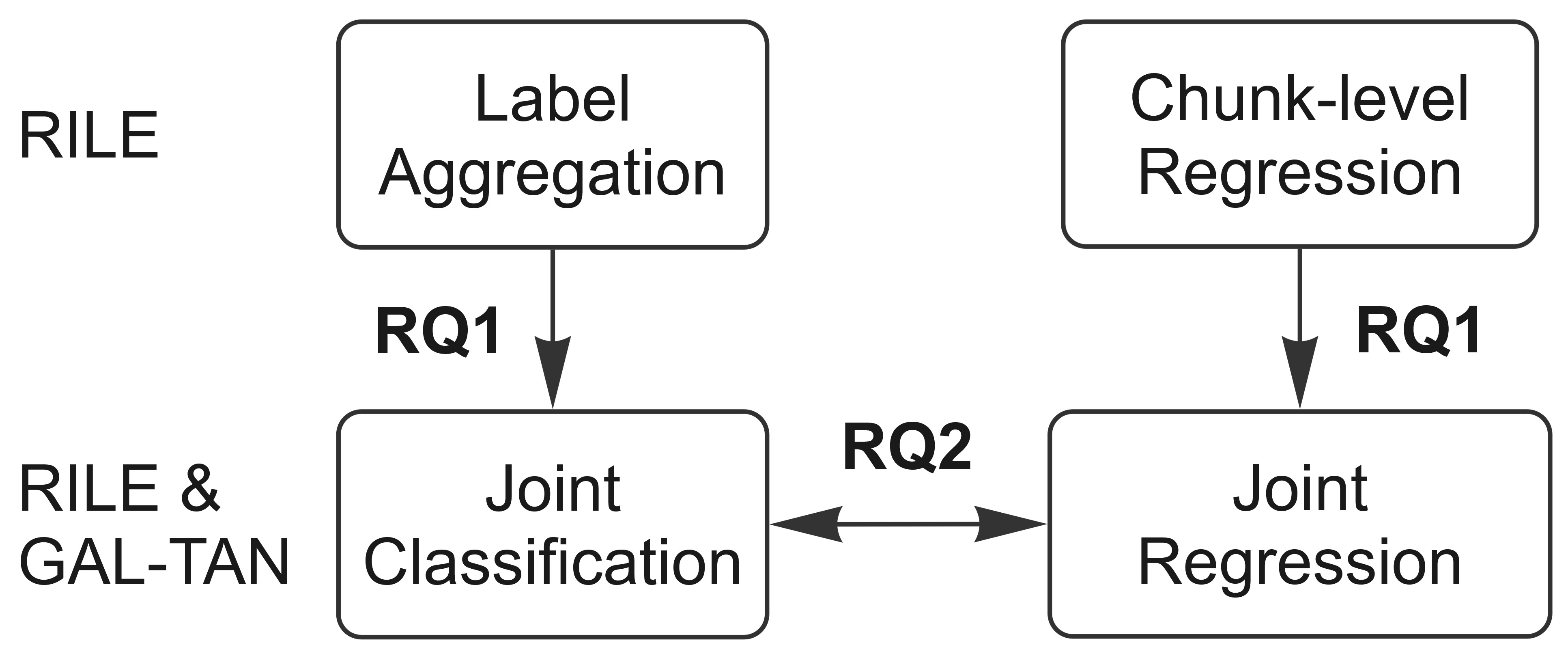}
    \caption{Experimental setup: Computational approaches to scaling and research questions.}
    \label{fig:RQs}
\end{figure}

\item[RQ2: How to deal with long input documents?] Traditional studies followed the MARPOR methodology in building a pipeline that first classifies individual sentences and then aggregates class frequencies into scale positions (\textit{label aggregation}).
\citet{nikolaev-etal-2023} also trained transformers to directly predict scale positions with a regression head, but still had to divide the input into smaller chunks for processing (\textit{chunk-level regression}). The trade-off between the two approaches remains underexplored, as well as the effect of the \textit{chunk size}, a crucial (hyper-)parameter.
\end{description}
Figure~\ref{fig:RQs} shows the resulting experimental design. We find for RQ1 that GAL-TAN can be predicted about as well as RILE, but that joint prediction does not improve performance. For RQ2, we establish that chunk size, somewhat surprisingly, matters rather little as far as performance is concerned, creating a continuum between regression and classification. Our code is available at \url{https://github.com/anna-golub/joint-rile-galtan}.


    

\section{Methods}

\subsection{Label Aggregation}
\label{sec:methods-label-aggr}
Mimicking a simplified version of the MARPOR coding scheme, label aggregation \cite{nikolaev-etal-2023} predicts the RILE score via sentence classification. A pre-trained Transformer encoder generates a sentence embedding to be fed to a classification head, which outputs one of the labels \textit{Right}, \textit{Left} or \textit{Other}. The model is trained with cross-entropy loss. 
Using Eq.~(\ref{eq:rile}), the sentence-level predictions are aggregated into manifesto-level positions between -1 (extreme left) and +1 (extreme right):
\begin{equation}\label{eq:rile}
     RILE = \frac{R - L}{R + L + O }
\end{equation}
where $R$,~$L$~and~$O$ are the number of sentences with the categories \textit{Right}, \textit{Left} and \textit{Other}, respectively. 

This approach can be extended to the GAL-TAN score estimation by replacing the RILE-specific categories with \textit{(G)AL}, \textit{(T)AN} and \textit{(O)ther}, and aggregating the predictions using the equation~(\ref{eq:galtan}). 

\begin{equation}\label{eq:galtan}
     GAL-TAN = \frac{G - T}{G + T + O }
\end{equation} 
The gold standard RILE and GAL-TAN categories are derived from fine-grained MARPOR sentence annotations in accordance with the literature \citep{volkens-rile-2013,MarksSteenbergen2004}. We refer to this method as \textit{label aggregation individual prediction}, as it estimates RILE and GAL-TAN independently from each other.

\subsection{Chunk-Level Regression}
Alternatively, the prediction tasks can be operationalized as regression, where a model maps a manifesto directly to a score in the range $[-1, 1]$ \cite{nikolaev-etal-2023}. While there is continuous research on robustly processing long inputs \citep{https://doi.org/10.1002/widm.70019}, the context window of the modern state-of-the-art Transformer encoders is nowhere near large enough to fit a whole manifesto. Nonetheless, some encoders allow inputs of up to 8192 tokens \cite{modernbert}, or over 650 MARPOR sentences, and therefore can approximate direct prediction with chunk-level processing. Specifically, the given manifesto is split into chunks of the maximum allowed length, after which a pre-trained encoder generates chunk embeddings, and a regression head trained on top outputs per-chunk scores. The model is trained with MSE loss, with the gold standard scores for a chunk calculated by aggregating the sentence labels (Eq.~\ref{eq:rile}) within the given chunk. For inference, the score of a whole manifesto is estimated as the average of the scores of its constituent chunks. Similarly to label aggregation, \textit{chunk-level regression individual prediction} can be employed equally to predict RILE and GAL-TAN scores.

\subsection{Joint Modeling}
\label{sec:joint}
We experiment with two strategies to move from individual to joint prediction (RQ1).

\paragraph{Multitask Training}
Among the multitask optimization methods such as multi-objective optimization, adversarial learning or neural architecture search, the one used most widely is \textit{scalarization}, where the joint loss is a linear combination of the losses for the individual tasks \citep{multitask-learning-book}. To apply scalarization to label aggregation and chunk-level regression, each pre-trained encoder is fine-tuned with two classification or regression heads on top, a RILE and a GAL-TAN specific one. The joint prediction loss function is the sum of the individual cross-entropy or MSE functions. The RILE and the GAL-TAN term are assigned equal weights because there is no implicit superiority of one over the other, and they are represented by the same number of data points. Because the RILE and GAL-TAN scores are correlated, multitask training is expected to improve over individual prediction, since the gradient of the encoder weights shows the direction where both the loss terms are minimized. 

\paragraph{Contrastive Training}

As an alternative to multitask training, we jointly train the label aggregation encoders with a contrastive objective. As our object function, we adopt \textit{triplet loss} \citep{schroff2015facenet}, used also to tune SBERT \citep{sentence-bert}. The training data is processed in triplets of an \textit{anchor concept} $a$, a \textit{positive concept} $p$ from the same class as $a$, and a \textit{negative concept} $n$ from a different class. Triplet loss pushes the similar instances $a$ and $p$ close together in the embedding space while driving the dissimilar $a$ and $n$ apart.
\begin{equation}
    max(||S_a - S_p|| - ||S_a - S_n|| + \epsilon, 0)
    \label{eq:triplet-loss}
\end{equation}
where $S_a, S_p, S_n$ are the vector representations of $a, p, n$; $||\cdot||$ is the distance metric. We employ the following two triplet mining strategies:

\begin{enumerate}
    \item RILE and GAL-TAN categories: The data points are selected randomly such that $a$~and~$p$ belong to the same RILE and the same GAL-TAN class while $n$ differs in at least one of the labels. This sampling strategy perpetuates the idea that the RILE and GAL-TAN categories are interdependent and each combination of them should occupy an isolated cluster in the sentence embedding space.

    \item Party: The triplets are chosen such that $a$ and $p$ are sentences from manifestos by the same party while $n$ is authored by a different one. When fine-tuned on the MARPOR sentences with party-based triplet loss, SBERT showed the best performance on pairwise party similarity estimation \citep{ceron-2022-dis-similarity}. Overall, party-based triplets allow the encoder to learn the general nature of manifesto text, which is helpful for positioning parties on a political axis.
\end{enumerate}
After contrastive tuning, the weights of the encoder are frozen and two classification heads are trained on top to predict the RILE and GAL-TAN categories. In addition, before the embeddings are fed to the classification heads, they are normalized using the \textit{whitening transformation} \citep{whitening}, which is known
to reduce the \textit{anisotropy} of the embedding and improve performance on various NLP tasks \citep{whitening}, including pairwise party similarity estimation \cite{ceron-2022-dis-similarity}.


\subsection{LLM Baseline}
\label{sec:llm-baseline}
Recent work \cite{lemens2025llms,ornstein2025llms,benoit2025llms} found that LLMs perform comparable to embedding-based approaches on political scaling. We therefore include an LLM in our experiment that implements the label aggregation approach, in order to  assess the relative performance of our embedding-based approaches.
We zero-shot instruct an LLM to annotate each sentence with one of the labels \textit{Right}, \textit{Left} and \textit{Other}, and estimate the manifesto RILE scores using Eq.~(\ref{eq:rile}). GAL-TAN individual prediction is approached analogously. We evaluate joint prediction capabilities by asking to assign both a RILE and a GAL-TAN category in one prompt. Since we only consider a single, simple experimental setting of this LLM with no prompt optimization (cf. Section \ref{sec:expt-setup}), we call this a 'baseline'.

\section{Experimental Setup}
\label{sec:expt-setup}
\subsection{Data}
\label{sec:data}
For our experiments, we adopt the \textsc{X-time} (\textsc{old-vs.-new)} setting from \citet{nikolaev-etal-2023}. This dataset primarily tests the generalizability of scaling models over time, from current to future election cycles, in countries seen during training. All models are trained on 1005 MARPOR manifestos for the years 2000--2018 (over 1M sentences). For each individual training setup, 10\% of this data is selected randomly and held out for validation. The available data for the years 2019--2023 (147 manifestos, 163K sentences), are used for testing. Data from before 2000 is excluded because of inconsistencies in the annotation. 

The model inputs are the translations of the original manifesto text created by \citet{nikolaev-etal-2023}\footnote{\url{https://osf.io/aypxd/overview}} with Opus-MT models available through EasyNMT.\footnote{\url{https://github.com/UKPLab/EasyNMT}} \citet{nikolaev-etal-2023} compared the use of MT with multilingual models and found almost identical results.

\subsection{Label Aggregation}

Following \citet{nikolaev-etal-2023}, 
we use a text encoder, here SBERT or ModernBERT, with a classification head on top. The joint models have two structurally identical heads, one per scale (cf. Section \ref{sec:joint}).

\paragraph{Classification head.}
The classification head is a multi-layer perceptron (MLP) consisting of two layers with the hidden size of 1024 and a tanh activation after the first layer. The output vector is passed through the softmax function to obtain a probability distribution.

\paragraph{SBERT} 
Sentence Transformers \cite{sentence-bert}, or SBERT, is a class of siamese Transformer models optimized for semantic textual similarity tasks. They have been shown to perform well on political party positioning, with SBERT label aggregation fine-tuned on the MARPOR data scoring the highest on RILE score prediction \cite{nikolaev-etal-2023}. Moreover, \citet{ceron-2022-dis-similarity} trained SBERT on MARPOR sentences for pairwise party similarity estimation, and after the whitening transformation, SBERT embeddings achieved the strongest correlation with the ground truth. For label aggregation, we use \texttt{all-mpnet-base-v2}, an SBERT model based on MPNet \cite{microsoft-mpnet} that is recommended as the general purpose Sentence Transformer with the highest embedding quality.\footnote{\url{https://sbert.net/docs/sentence_transformer/pretrained\_models.html\#original-models}} However, due to its small context size, it is not suitable as an encoder for chunk-level regression.

\paragraph{ModernBERT}
ModernBERT \cite{modernbert} is a recent update on the original BERT model \citep{devlin-etal-2019-bert}. It incorporates rotary positional embeddings (RoPE), pre-normalization blocks and an updated activation function, all established improvements on the original Transformer architecture \citep{modernbert}. The self-attention mechanism at the core of the Transformer architecture is associated with a quadratic computational cost, which ModernBERT addresses by only using \textit{global self-attention} in every third layer. The rest are \textit{local attention} layers, where each input token only attends to the tokens close to it. ModernBERT outperforms BERT variants of a similar size and is competitive with bigger, slower variants such as GTE-en-MLM \citep{gte-en-mlm} and DeBERTa-v3-large \citep{deberta}.
We use ModernBERT-base in our experiments.\footnote{\url{https://huggingface.co/answerdotai/ModernBERT-base}}

\begin{table*}[tb!]
\centering
\begin{tabular}{@{}llcccccc@{}}
\toprule
 &  & \multicolumn{2}{c}{Individual} & \multicolumn{2}{c}{Joint Multitask} & \multicolumn{2}{c}{Joint Contrastive} \\ 
 &  & RILE & GAL-TAN & RILE & GAL-TAN & RILE & GAL-TAN \\ 
 \cmidrule(l){3-4} 
 \cmidrule(lr){5-6} 
 \cmidrule(r){7-8} 
\multirow{4}{*}{Label Aggregation} & 
 SBERT [N23] & \textbf{0.88} & - & - & - & - & - \\
& SBERT (ours) & \textbf{0.88} & \textbf{0.83} & \textbf{0.87} & \textbf{0.84} & 0.83 & 0.77 \\
 & ModernBERT & \textbf{0.87} & \textbf{0.83} & \textbf{0.86} & \textbf{0.83} & 0.80 & 0.74 \\
  & LLM (Olmo 3) & 0.67 & 0.53 & 0.61 & 0.48 & -  & - \\
 \cmidrule(l){3-4} 
 \cmidrule(lr){5-6} 
 \cmidrule(r){7-8} 
\multirow{3}{*}{Chunk Regression} & 
BigBird [N23] & 0.71 & - & - & - & - & -  \\
& BigBird (ours) & 0.84 & \textbf{0.84} & 0.84 & 0.82 & - & - \\
 & ModernBERT & 0.79 & 0.78 & 0.83 & 0.79 & - & - \\ 
 \bottomrule
\end{tabular}
\caption{Rank correlation with the gold standard manifesto scores. The best performance on RILE and on GAL-TAN each is highlighted in bold. When multiple values are boldfaced for a given target, the difference between them is not statistically significant. The joint contrastive models are SBERT-RILE-GAL-TAN\textsubscript{whiten} and ModernBERT-RILE-GAL-TAN. [N23] is \citet{nikolaev-etal-2023}.}
\label{tab:results-main}
\end{table*}

\subsection{Chunk-Level Regression}

For chunk-level regression, we combine, again, a text encoder (BigBird and ModernBERT) with a regression head. As before, the joint versions have two structurally identical heads.

\paragraph{Regression head.} The regression head is the same MLP described above, except the final softmax layer is replaced with a single tanh-activated unit to obtain predictions in the $-1\dots1$ range.

\paragraph{BigBird.}
\citet{bigbird} tackled the complexity of the self-attention computation by approximating it with \textit{sparse attention}, which prunes the set of possible attention links to achieve a speed-up. The resulting model can deal with a context window of 4096 tokens. BigBird was initialized from RoBERTa \citep{roberta} and further trained on a large web corpus. When evaluated on long-input question answering and long document classification, BigBird set the new state of the art on several datasets and otherwise demonstrated competitive performance. In \citet{nikolaev-etal-2023}, BigBird was the best out of the evaluated long-input encoder but only 
scored moderately well compared to the label aggregation setup. 
BigBird chunks consist of 173 sentences on average, making the mean manifesto length 6--7 chunks. We use BigBird-base in our experiments.\footnote{\url{https://huggingface.co/google/bigbird-roberta-base}}

\paragraph{ModernBERT.}
ModernBERT employs Flash Attention \citep{flashattention} to extend the original BERT context window of 512 tokens to 8192, which makes it suitable also for long input regression. On average, one chunk of 8192 tokens fits a maximum of 315 MARPOR sentences, and the mean manifesto length is 3-4 chunks.
Hence, ModernBERT is expected to perform well as the encoder in both label aggregation and chunk-level regression, allowing for a direct comparison between the approaches on a conceptual level.

\subsection{LLM Baseline} 

 We employ
\texttt{Olmo-3-7B-Instruct} \cite{olmo2026olmo3}, one of the few LLMs with a completely open training procedure \citep{LiesenfeldDingemanse2024}, in a zero-shot setting. See Appendix~\ref{app:exp-setup} for the prompts.

\subsection{Evaluation}

To make our results comparable to \citet{nikolaev-etal-2023}, we
measure the RILE and GAL-TAN estimation quality as Spearman rank correlation coefficient between the predicted and the gold standard scores at the manifesto level. This metric evaluates the ability of the models to correctly rank parties on the scales, rather than predict absolute positions.

We train each supervised model with 5 random seeds and report the averages of the per-seed evaluation metrics. For the LLM baseline, we sample 5 responses per data point and average the evaluation metrics over the LLM answers. For all pairs of models with $\leq$15 percentage points of difference in performance, we test the statistical significance of the difference by running the bootstrap resampling test with $10,000$ resamples and a $95\%$ confidence interval \citep{bootstrap}. 

\section{Results}

\begin{figure*}[tb!]
    \centering
    \begin{subfigure}{}
    \centering
    \includegraphics[width=0.4\linewidth]{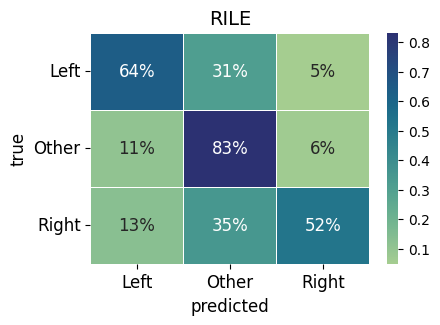}
    \label{fig:heatmap-rile-ind}
    \end{subfigure}
    \begin{subfigure}{}
    \centering
    \includegraphics[width=0.57\linewidth]{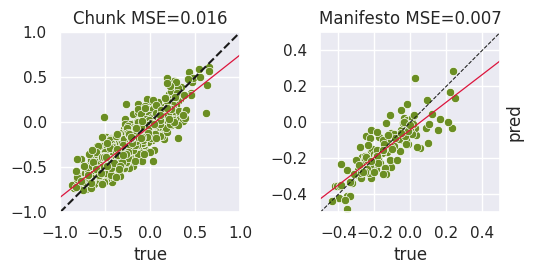}
    \label{fig:scatterplot-bigbird}
    \end{subfigure}
    \caption{\textit{Left}: Confusion matrix heatmap: Label aggregation RILE individual prediction with SBERT. \textit{Right}: Predicted vs. true scatterplot: Chunk-level RILE regression with BigBird (test set, random seed 7).}
    \label{fig:error-analysis}
\end{figure*}

\subsection{Individual Prediction}
\label{sec:results-ind}
The main results are reported in Table~\ref{tab:results-main}.
Label aggregation individual prediction with SBERT achieves a very strong correlation of $\rho=0.88$ on RILE, a very close replication of the result in \citet{nikolaev-etal-2023}.

The GAL-TAN scale is slightly more difficult to predict, even if not by a large margin, with the best results around $\rho=0.83$ to $0.84$. This may be related to the more complex nature of cultural, as opposed to economic, stances \citep{Kurella16042026}, as well as the training data imbalance that is more severe for GAL-TAN than for RILE.

Comparing the different implementations of label aggregation, we note that ModernBERT performs on a par with SBERT on both scales. Zero-shot prediction with Olmo 3 is much less robust than the supervised methods with $\rho=0.67/0.53$. While it might be possible to improve on these results with the usual bag of LLM improvement strategies (prompt engineering, few-shot setup, larger models, chain-of-thought reasoning), this is outside the scope of our study. We treat this result mostly as illustration that the embedding-based approaches we consider are not immediately outclassed by LLM-based approaches. We note that \citet{benoit2025llms} report  (linear) correlation scores of a similar magnitude in an experiment with a considerably more complex LLM setup on  MARPOR data ($r=0.57$ on \textit{Taxes vs. Spending} and $r=0.68$ on the \textit{Social} axis).\footnote{From the supplementary materials of \citet{benoit2025llms}.}

For chunk-level regression, prediction with BigBird is on a par with label aggregation on GAL-TAN and slightly behind on RILE with $\rho=0.84$ on both. This is a new qualitative finding, as chunk-level regression with BigBird was reported by \citet{nikolaev-etal-2023} to fall behind label aggregation. 
ModernBERT underperforms BigBird somewhat; however, this may be a consequence of our choice to use the maximum chunk size (twice as high for ModernBERT as for BigBird), cf. Section~\ref{sec:exp-chunk-size}.

\paragraph{Error Analysis.}
While the manifesto-level rank correlation is strong, there is still space for improvement at the model output level. Due to the skew towards the label \textit{Other} in the training data, label aggregation misclassifies over $30\%$ of the sentences marked \textit{Right}, \textit{Left}, \textit{GAL} and \textit{TAN} (see Figure~\ref{fig:error-analysis}, left). The aggregation of the sentence-level predictions according to Eq.~(\ref{eq:rile}) and (\ref{eq:galtan}) respectively appears to smooth out the errors, but the resulting manifesto scores suffer from regression to the mean (i.e. zero): They are correct in sign but too small in magnitude. Figure~\ref{fig:error-analysis} (right) shows that  chunk-level regression shows the same effects, which can be interpreted as low model confidence.

\subsection{Joint Prediction (RQ1)}
Contrary to our expectations from RQ\ 1, joint multitask training does not improve over individual prediction in almost any setting. In the label aggregation setting, joint multitask learning does not have any major effect on the embedding-based model. Prompting Olmo to output labels for both scales jointly even has a clear negative effect. 

Contrastive tuning also impedes the quality of the predictions throughout. 
As shown in Appendix~\ref{app:results}, performance drops further when selecting triplets based on party and while, in line with \citet{ceron-2022-dis-similarity}, the whitening transformation has an overall positive effect on SBERT, it is detrimental to ModernBERT.

Finally, in the chunk regression setting, joint multitask training leads to slightly worse results for BigBird. While ModernBERT improves slightly with joint multitask training, it still scores lower than BigBird due to its lower starting point.


\begin{figure}[tb!]
    \centering
    \includegraphics[width=\linewidth]{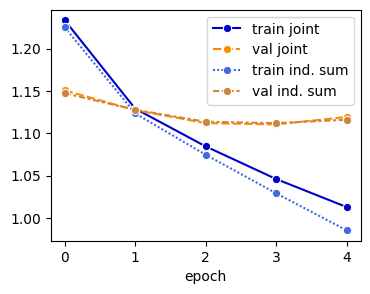}
    \caption{SBERT learning curves: joint multitask learning vs. the sum of the RILE and GAL-TAN losses during individual optimization (random seed 7).}
    \label{fig:learn-curve-sbert-ind-sum-vs-joint}
\end{figure}

\paragraph{Training dynamics.} To better understand the consistent failure of our models to profit from joint prediction, we analyze the behavior of the RILE and GAL-TAN train and validation losses during training. As Figure~\ref{fig:learn-curve-sbert-ind-sum-vs-joint} shows, the losses behave in the same way as the sum of the losses computed during individual optimization. Therefore, the two objectives appear to neither sabotage nor support each other and are effectively learned separately. A similar effect is observed for chunk-level regression. One underlying reason for this might be that the correlation between RILE and GAL-TAN in the training data ($\rho=0.48$) is still too weak for the models to pick up and capitalize on. Some category combinations, e.g., \textit{Right}+\textit{GAL}, are underrepresented in the training data (see Appendix~\ref{app:exp-setup}), so learning to predict them might be challenging.

\paragraph{Ceiling Effect.} Another potential explanation of the models' failure to improve is a ceiling effect, i.e., the individual predictions are already bounded by the reliability of the data so that better modeling mechanisms cannot further improve the results.
Indeed, a MARPOR coder reliability study 
\cite[Table 3]{coder_reliability_2012}
reports agreement numbers that amount to a macro-F1 score of $0.66$ between two human annotators that classify sentences into the RILE categories \textit{Right}, \textit{Left} and \textit{Other}. This is coincidentally the exact performance achieved by SBERT on the classification task (cf. Appendix~\ref{app:results}). On GAL-TAN, the model's macro-F1 is $0.67$ whereas the human estimate is $0.61$. We acknowledge that this constitutes only partial evidence: The analysis by \citet{coder_reliability_2012} was carried out on a sample of about 1700 English-language text units. However, this is, to our knowledge, the only MARPOR coder reliability study that we can establish a numerical comparison with.
That being said, a ceiling effect remains a plausible interpretation.


\subsection{Long Input Documents (RQ2)}
\label{sec:exp-chunk-size}

A notable result from Table~\ref{tab:results-main} is that ModernBERT, while suitable for both label aggregation-based and regression-based prediction (cf.~\S\ref{sec:expt-setup}), performs substantially better on label aggregation by a margin of 5-8 points in rank correlation. This may be due to the model becoming less robust with longer inputs, or due to the difference between classification and regression as tasks. To investigate the trade-offs between the two task formulations, we vary the chunk sizes for joint ModernBERT-based chunk-level regression on a logarithmic scale from $n$=1 to around $n\approx300$ sentences on average, covering the full range of the possible input length (see Appendix~\ref{app:exp-setup} for details). Each model is trained once with the random seed 7.

\begin{figure}[tb!]
        \centering
        \includegraphics[width=\linewidth]{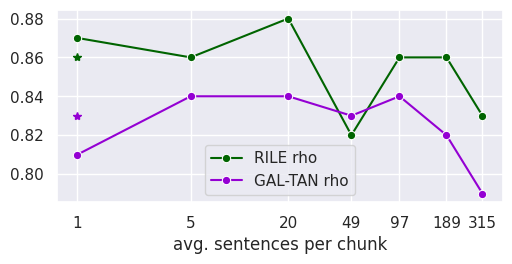}
    \caption{Joint chunk-level regression with ModernBERT on various chunk sizes (test set, random seed 7). The dots are connected for readability only. The stars mark the performance of joint label aggregation with ModernBERT. 
    }
    \label{fig:chunk-size}
\end{figure}

Strikingly, the models score in the same range \textit{regardless of the chunk size}, namely $\rho$=0.82--0.87 on RILE and $\rho$=0.79--0.84 on GAL-TAN (see Figure~\ref{fig:chunk-size}\footnote{The results in Table\ \ref{tab:results-main} adopt the largest possible average chunk size ($n$=315 sentences).}). The bootstrap test shows that all the models are statistically on a par. With our current setup, we however cannot rule out the possibility that the low variance is due to our training with a single random seed. 

\begin{figure*}[tb!]
    \centering
    \includegraphics[width=\linewidth]{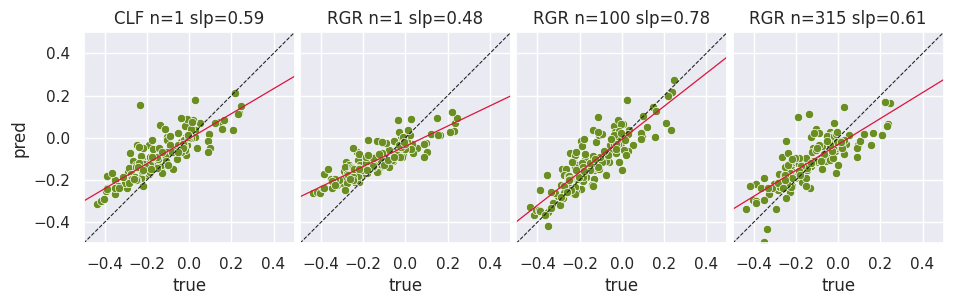}
    \caption{Classification-regression continuum: RILE true vs. predicted manifesto scores scatterplot (test set, random seed 7). $n$ is the number of input sentences; $slp$ is the slope of the regression line (red). Lower $slp$ indicates stronger regression to the mean.}
    \label{fig:clf-rgr-continuum}
\end{figure*}

\paragraph{Classification-Regression Continuum.}
Note that for the chunk size $n=1$ sentence, the regression model is trained to predict $-1$, $0$, or $-1$, depending on the class of the sentence. This is very similar to the training task of the label aggregation setup, only with a regression instead of a classification head. Thus, ModernBERT can be seen as supporting a continuum between classification and regression.
This does not mean that the results are exactly the same, though: Regression with chunk size $n$=1 is numerically on a par with label aggregation on RILE (0.87 vs. 0.86) but slightly worse on GAL-TAN (0.81 vs. 0.83). Indeed, the regression is less precise on just $n$=1 sentence per chunk than on max. $n$=100 (97 on average): MSE $=0.23/0.15$ (RILE/GAL-TAN) for $n$=1 vs. MSE $=0.022/0.016$ for $n$=100 at the chunk level. 
As discussed in Section~\ref{sec:results-ind}, performance levels out once the chunk estimates are averaged to represent full manifestos, producing MSE $=0.009/0.007$ for $n$=1 and MSE $=0.006/0.006$ for $n$=100 at the manifesto level.

Another drawback of small chunk sizes is a particularly strong regression to the mean. In Figure\ \ref{fig:clf-rgr-continuum} the predicted manifesto-level RILE scores are plotted against the gold standard (the picture is the same for GAL-TAN). In these scatterplots, ideal predictions would lie along a regression line with slope 1, but in reality, the slopes are below 1, meaning both classification and regression predictions are too small in magnitude.
Regression with chunk size $n$=100 obtains the overall highest slope of 0.78, indicating that this model more accurately approximates not only the gold standard ranking but also the absolute values of positions.
For very large chunk sizes, both correlation and slope decrease, showing that models still struggle to extract information reliably from very large contexts \citep{3780338.3783035}.



Taken together, these results indicate that regression-based direct prediction of positions on political scales provides an alternative to label aggregation. 
As the largest chunk sizes lead to a small drop in performance, and very small sizes suffer from regression to the mean, medium context sizes (20-100 sentences) are the most robust choice.


\section{Related Work}

Early work on automating political scaling relied on word frequencies for supervised \cite{laver2003-wordscores} and unsupervised \cite{slapin2008-wordfish} estimation of party positions. Once NLP shifted to distributional semantics, \citet{glavas-etal-2017} employed word embeddings for unsupervised analysis of multilingual data, computing pairwise party similarities and rescaling them to obtain positions on an axis. This line of work was continued by \citet{Rheault_Cochrane_party2vec} and \citet{glavas-nanni-semscale}.

More recently, encoder-only Transformer models have been used for pairwise party similarity estimation, overall \cite{ceron-2022-dis-similarity} and within policy domains with the aid of a sentence domain classifier \cite{ceron-etal-2023-additive}. \citet{dayanik-etal-2022-improving} fine-tuned BERT to predict fine-grained MARPOR sentence labels, improving the performance on infrequent classes with hierarchical classification.

The success of LLMs on various NLP tasks \cite{llm-survey-2025} has called for research on their applicability to political scaling. \citet{lemens2025llms} queried state-of-the-art closed-source LLMs to map a given sentence to a position on a economic or social axis, achieving strong correlation with the expert and crowd-sourced gold standard. \citet{benoit2025llms} ensembled the predictions of several zero- and few-shot closed-source LLMs which were prompted to summarize long inputs first and then scale them. The results show strong correlation with expert positions but only moderate correlation with the MARPOR ground truth, as discussed in~\S\ref{sec:results-ind}.

\section{Conclusions}

In this paper, we have systematically evaluated embedding-based approaches to party positioning based on election manifestos. Our study was carried out on a sample from the MARPOR corpus, evaluating generalization from past to future election cycles. We focused on comparing two approaches, label aggregation and chunk-level regression, and established two main findings:

First, joint estimation of party positions on the RILE and GAL-TAN scales does not improve prediction quality. While this is at first glance a disappointing result, the performance of our non-joint models already approaches a plausible ceiling arising from inter-annotator disagreement. It would be worthwhile, in future work, to test joint modeling approaches in more challenging scenarios, such as generalization to new countries \citep{nikolaev-etal-2023} or party position prediction based on less data -- modeling situations where domain knowledge tends to help \citep{dayanik-etal-2022-improving}.

Second, chunk-level regression with ModernBERT offers a viable alternative to label aggregation in political positioning, in particular when medium chunk sizes (20-100 sentences) are chosen. This result is at the same time encouraging and disappointing: while it is conceptually more elegant to directly predict positions without an intermediate labeling step, and the regression approach models the actual score distribution better than the classification approach, the decrease in performance on the longest chunks indicates that manifestos are still too long for current Transformer LMs to obtain good end-to-end learning results. Interestingly, chunk-level regression models do not require gold standard annotation at the sentence level for training. In future work, the detailed MARPOR annotation could potentially be replaced by a continuous position annotation at the chunk level, possibly framed as a ranking task \citep{CARLSON_MONTGOMERY_2017}.

\section*{Limitations}

Our study only investigated a single, comparatively simple setup for political party positioning: generalizing from previous to future election cycles. The RILE and GAL-TAN scales themselves have been criticised for being inflexible and overly simplistic, as they are rooted in saliency theory and use the same codebook for all countries and time periods \cite{albright2010,jahn2023}. 

We relied on the quality of the MT system used by \citet{nikolaev-etal-2023} and did not experiment with multilingual models\footnote{A verified English translation of the MARPOR data has since become available. \url{https://manifesto-project.wzb.eu/information/documents/translation}}. Due to limited resources, we used the smaller versions of the pre-trained models, and the hyperparameter search was performed manually. The LLM approach that we included in our experiments was not optimized regarding its prompt, nor did we set up a few-shot variant; in this sense, it can be considered an unsupervised (or semi-supervised) point of comparison for the fully supervised models that we focused on.

\bibliography{custom}

@inproceedings{Brigevich2017-GALTAN,
  title={{Unpacking the social (GAL/TAN) dimension of party politics: Euroscepticism and party positioning on Europe’s “other”}},
  author={Anna Brigevich and William B. Smith},
  year={2017},
  url={https://api.semanticscholar.org/CorpusID:211235760},
  booktitle={Proceedings of the European Union Studies Association Biannual Conference},
  address={Miami}
}

@article{Kurella16042026,
author = {Anna-Sophie Kurella and Milena Rapp},
title = {Unfolding {GAL-TAN}: the multi-dimensional nature of public opinion in Western Europe},
journal = {West European Politics},
volume = {49},
number = {3},
pages = {841--866},
year = {2026},
publisher = {Routledge},
doi = {10.1080/01402382.2025.2466117}}

@InProceedings{LiesenfeldDingemanse2024,
  author    = {Liesenfeld, Andreas and Dingemanse, Mark},
  title     = {Rethinking open source generative {AI}: open-washing and the {EU AI Act}},
  booktitle = {Proceedings of the ACM Conference on Fairness, Accountability, and Transparency},
  year      = 2024,
  pages     = {1774–1787},
  isbn      = 9798400704505,
  url       = {https://doi.org/10.1145/3630106.3659005},
  doi       = {10.1145/3630106.3659005},
  numpages  = 14,
  address  = {Rio de Janeiro, Brazil},
}

@article{https://doi.org/10.1002/widm.70019,
author = {Alva Principe, Renzo and Chiarini, Nicola and Viviani, Marco},
title = {Long Document Classification in the Transformer Era: A Survey on Challenges, Advances, and Open Issues},
journal = {WIREs Data Mining and Knowledge Discovery},
volume = {15},
number = {2},
pages = {e70019},
url = {https://doi.org/10.1002/widm.70019},
year = {2025}
}

@article{koedam_binding_steenbergen_2025,
  author    = {Koedam, Jelle and Binding, Garret and Steenbergen, Marco R.},
  title     = {Multidimensional Party Polarization in {Europe}: Cross-Cutting Divides and Effective Dimensionality},
  journal   = {British Journal of Political Science},
  volume    = {55},
  pages     = {e24},
  year      = {2025},
  publisher = {Cambridge University Press},
  url       = {https://doi.org/10.1017/S0007123424000474},
}

@article{ROVNY2025102981,
title = {The 2024 {Chapel Hill} Expert Survey on political party positioning in {E}urope: Twenty-five years of party positional data},
journal = {Electoral Studies},
volume = {97},
pages = {102981},
year = {2025},
issn = {0261-3794},
url = {https://doi.org/10.1016/j.electstud.2025.102981},
author = {Jan Rovny and Jonathan Polk and Ryan Bakker and Liesbet Hooghe and Seth Jolly and Gary Marks and Marco Steenbergen and Milada Anna Vachudova}
}

@book{benoit_laver_ppmd_2006,
author = {Benoit, Kenneth and Laver, Michael},
year = {2006},
title = {Party Policy in Modern Democracies},
isbn = {9780203028179},
publisher = {Routledge},
doi = {10.4324/9780203028179},
address={London}
}

@techreport{rile_budge_2013,
    author = {Budge, Ian},
    title = {The Standard Left-Right Scale},
    institution = {Comparative Manifesto Project},
    year = {2013},
    url = {https://manifesto-project.wzb.eu/down/papers/budge_right-left-scale.pdf}
}

@incollection{volkens-rile-2013,
    author = {Volkens, Andrea and Bara, Judith and Budge, Ian and McDonald, Michael D. and Best, Robin and Franzmann, Simon},
    isbn = {9780199640041},
    title = {Understanding and Validating the Left-Right Scale {(RILE)}},
    booktitle = {Mapping Policy Preferences From Texts: Statistical Solutions for Manifesto Analysts},
    publisher = {Oxford University Press},
    year = {2013},
    month = {11},
    doi = {10.1093/acprof:oso/9780199640041.003.0006},
    url = {https://doi.org/10.1093/acprof:oso/9780199640041.003.0006},
    eprint = {https://academic.oup.com/book/0/chapter/197751501/chapter-ag-pdf/44604447/book_27649_section_197751501.ag.pdf},
}

@article{multitask-learning-book,
	author = {Yu, Jun and Liu, Xiaokang and Luo, Chongliang and Zhou, Rong and Liu, Yixin and Hu, Jie and Chen, Jianmin and Zhang, Ke and Zhang, Dazheng and Shen, Yishan and Adhikarla, Eashan and Dai, Yutong and Zhang, Kai and Kong, Zhaoming and Ye, Wenxuan and Yin, Yilong and Namboodiri, Vinod and Davison, Brian and Moore, Jason and Chen, Yong},
	journal = {Harvard Data Science Review},
	number = {3},
	year = {2025},
	url = {https://hdsr.mitpress.mit.edu/pub/7fcc3jhv},
	publisher = {The MIT Press},
	title = {{Multitask Learning 1997--2024: Part II Regularization and Optimization}},
	volume = {7},
}

@inproceedings{schroff2015facenet,
  title={Facenet: A unified embedding for face recognition and clustering},
  author={Schroff, Florian and Kalenichenko, Dmitry and Philbin, James},
  url = {https://doi.ieeecomputersociety.org/10.1109/CVPR.2015.7298682},
  booktitle={Proceedings of the IEEE conference on computer vision and pattern recognition},
  pages={815--823},
  year={2015}
}

@inproceedings{devlin-etal-2019-bert,
    title = "{BERT}: Pre-training of Deep Bidirectional Transformers for Language Understanding",
    author = "Devlin, Jacob  and
      Chang, Ming-Wei  and
      Lee, Kenton  and
      Toutanova, Kristina",
    editor = "Burstein, Jill  and
      Doran, Christy  and
      Solorio, Thamar",
    booktitle = "Proceedings of the 2019 Conference of the North {A}merican Chapter of the Association for Computational Linguistics: Human Language Technologies, Volume 1 (Long and Short Papers)",
    month = jun,
    year = "2019",
    address = "Minneapolis, Minnesota",
    publisher = "Association for Computational Linguistics",
    url = "https://aclanthology.org/N19-1423/",
    doi = "10.18653/v1/N19-1423",
    pages = "4171--4186",
}

@inproceedings{nikolaev-etal-2023,
    title = "Multilingual estimation of political-party positioning: From label aggregation to long-input Transformers",
    author = "Nikolaev, Dmitry  and
      Ceron, Tanise  and
      Pad{\'o}, Sebastian",
    editor = "Bouamor, Houda  and
      Pino, Juan  and
      Bali, Kalika",
    booktitle = "Proceedings of the 2023 Conference on Empirical Methods in Natural Language Processing",
    month = dec,
    year = "2023",
    address = "Singapore",
    publisher = "Association for Computational Linguistics",
    url = "https://aclanthology.org/2023.emnlp-main.591/",
    doi = "10.18653/v1/2023.emnlp-main.591",
    pages = "9497--9511",
}

@article{laver2003-wordscores,
  title={Extracting policy positions from political texts using words as data},
  author={Laver, Michael and Benoit, Kenneth and Garry, John},
  journal={American political science review},
  volume={97},
  url = {https://doi.org/10.1017/S0003055403000698},
  number={2},
  pages={311--331},
  year={2003},
  publisher={Cambridge University Press}
}

@article{slapin2008-wordfish,
  title={A scaling model for estimating time-series party positions from texts},
  author={Slapin, Jonathan B and Proksch, Sven-Oliver},
  journal={American Journal of Political Science},
  volume={52},
  url = {https://doi.org/10.1111/j.1540-5907.2008.00338.x},
  number={3},
  pages={705--722},
  year={2008},
  publisher={Wiley Online Library}
}

@inproceedings{ceron-etal-2023-additive,
    title = "Additive manifesto decomposition: A policy domain aware method for understanding party positioning",
    author = "Ceron, Tanise  and
      Nikolaev, Dmitry  and
      Pad{\'o}, Sebastian",
    editor = "Rogers, Anna  and
      Boyd-Graber, Jordan  and
      Okazaki, Naoaki",
    booktitle = "Findings of the Association for Computational Linguistics: ACL 2023",
    month = jul,
    year = "2023",
    address = "Toronto, Canada",
    publisher = "Association for Computational Linguistics",
    url = "https://aclanthology.org/2023.findings-acl.499/",
    doi = "10.18653/v1/2023.findings-acl.499",
    pages = "7874--7890",
}

@inproceedings{modernbert,
    title = "Smarter, Better, Faster, Longer: A Modern Bidirectional Encoder for Fast, Memory Efficient, and Long Context Finetuning and Inference",
    author = {Warner, Benjamin  and
      Chaffin, Antoine  and
      Clavi{\'e}, Benjamin  and
      Weller, Orion  and
      Hallstr{\"o}m, Oskar  and
      Taghadouini, Said  and
      Gallagher, Alexis  and
      Biswas, Raja  and
      Ladhak, Faisal  and
      Aarsen, Tom  and
      Adams, Griffin Thomas  and
      Howard, Jeremy  and
      Poli, Iacopo},
    editor = "Che, Wanxiang  and
      Nabende, Joyce  and
      Shutova, Ekaterina  and
      Pilehvar, Mohammad Taher",
    booktitle = "Proceedings of the 63rd Annual Meeting of the Association for Computational Linguistics (Volume 1: Long Papers)",
    month = jul,
    year = "2025",
    address = "Vienna, Austria",
    publisher = "Association for Computational Linguistics",
    url = "https://aclanthology.org/2025.acl-long.127/",
    doi = "10.18653/v1/2025.acl-long.127",
    pages = "2526--2547",
    ISBN = "979-8-89176-251-0",
}

@inproceedings{bigbird,
  title={Big bird: Transformers for longer sequences},
  url = {https://papers.neurips.cc/paper_files/paper/2020/file/c8512d142a2d849725f31a9a7a361ab9-Paper.pdf},
  author={Zaheer, Manzil and Guruganesh, Guru and Dubey, Kumar Avinava and Ainslie, Joshua and Alberti, Chris and Ontanon, Santiago and Pham, Philip and Ravula, Anirudh and Wang, Qifan and Yang, Li and others},
  booktitle={Advances in neural information processing systems},
  pages={17283--17297},
  year={2020}
}

@inproceedings{sentence-bert,
    title = "Sentence-{BERT}: Sentence Embeddings using {S}iamese {BERT}-Networks",
    author = "Reimers, Nils  and
      Gurevych, Iryna",
    editor = "Inui, Kentaro  and
      Jiang, Jing  and
      Ng, Vincent  and
      Wan, Xiaojun",
    booktitle = "Proceedings of the 2019 Conference on Empirical Methods in Natural Language Processing and the 9th International Joint Conference on Natural Language Processing (EMNLP-IJCNLP)",
    month = nov,
    year = "2019",
    address = "Hong Kong, China",
    publisher = "Association for Computational Linguistics",
    url = "https://aclanthology.org/D19-1410/",
    doi = "10.18653/v1/D19-1410",
    pages = "3982--3992"
}

@misc{llm-survey-2025,
      title={{Large Language Models: A Survey}}, 
      author={Shervin Minaee and Tomas Mikolov and Narjes Nikzad and Meysam Chenaghlu and Richard Socher and Xavier Amatriain and Jianfeng Gao},
      year={2025},
      eprint={2402.06196},
      archivePrefix={arXiv},
      primaryClass={cs.CL},
      url={https://arxiv.org/abs/2402.06196}, 
}

@inproceedings{glavas-etal-2017,
    title = "Unsupervised Cross-Lingual Scaling of Political Texts",
    author = "Glava{\v{s}}, Goran  and
      Nanni, Federico  and
      Ponzetto, Simone Paolo",
    editor = "Lapata, Mirella  and
      Blunsom, Phil  and
      Koller, Alexander",
    booktitle = "Proceedings of the 15th Conference of the {E}uropean Chapter of the Association for Computational Linguistics: Volume 2, Short Papers",
    month = apr,
    year = "2017",
    address = "Valencia, Spain",
    publisher = "Association for Computational Linguistics",
    url = "https://aclanthology.org/E17-2109/",
    pages = "688--693"
}

@article{Rheault_Cochrane_party2vec, 
title={Word Embeddings for the Analysis of Ideological Placement in Parliamentary Corpora}, 
volume={28}, 
DOI={10.1017/pan.2019.26}, 
number={1}, 
journal={Political Analysis}, 
author={Rheault, Ludovic and Cochrane, Christopher}, 
year={2020}, 
pages={112–133}
}

@article{glavas-nanni-semscale,
author = {Nanni, Federico and Glava\v{s}, Goran and Rehbein, Ines and Ponzetto, Simone Paolo and Stuckenschmidt, Heiner},
title = {Political Text Scaling Meets Computational Semantics},
year = {2022},
issue_date = {November 2021},
publisher = {Association for Computing Machinery},
address = {New York, NY, USA},
volume = {2},
number = {4},
issn = {2691-1922},
url = {https://doi.org/10.1145/3485666},
doi = {10.1145/3485666},
journal = {ACM/IMS Trans. Data Sci.},
month = may,
articleno = {29},
numpages = {27}
}

@inproceedings{ceron-2022-dis-similarity,
    title = "Optimizing text representations to capture (dis)similarity between political parties",
    author = "Ceron, Tanise  and
      Blokker, Nico  and
      Pad{\'o}, Sebastian",
    editor = "Fokkens, Antske  and
      Srikumar, Vivek",
    booktitle = "Proceedings of the 26th Conference on Computational Natural Language Learning (CoNLL)",
    month = dec,
    year = "2022",
    address = "Abu Dhabi, United Arab Emirates (Hybrid)",
    publisher = "Association for Computational Linguistics",
    url = "https://aclanthology.org/2022.conll-1.22/",
    doi = "10.18653/v1/2022.conll-1.22",
    pages = "325--338"
}

@misc{whitening,
      title={Whitening Sentence Representations for Better Semantics and Faster Retrieval}, 
      author={Jianlin Su and Jiarun Cao and Weijie Liu and Yangyiwen Ou},
      year={2021},
      eprint={2103.15316},
      archivePrefix={arXiv},
      primaryClass={cs.CL},
      url={https://arxiv.org/abs/2103.15316}, 
}

@inproceedings{
adamw,
title={Decoupled Weight Decay Regularization},
author={Ilya Loshchilov and Frank Hutter},
booktitle={International Conference on Learning Representations},
year={2019},
url={https://openreview.net/forum?id=Bkg6RiCqY7},
}

@inproceedings{microsoft-mpnet,
 author = {Song, Kaitao and Tan, Xu and Qin, Tao and Lu, Jianfeng and Liu, Tie-Yan},
 booktitle = {Advances in Neural Information Processing Systems},
 editor = {H. Larochelle and M. Ranzato and R. Hadsell and M.F. Balcan and H. Lin},
 pages = {16857--16867},
 publisher = {Curran Associates, Inc.},
 title = {MPNet: Masked and Permuted Pre-training for Language Understanding},
 url = {https://proceedings.neurips.cc/paper_files/paper/2020/file/c3a690be93aa602ee2dc0ccab5b7b67e-Paper.pdf},
 volume = {33},
 year = {2020}
}

@inproceedings{roberta,
    title = "A Robustly Optimized {BERT} Pre-training Approach with Post-training",
    author = "Zhuang, Liu  and
      Wayne, Lin  and
      Ya, Shi  and
      Jun, Zhao",
    editor = "Li, Sheng  and
      Sun, Maosong  and
      Liu, Yang  and
      Wu, Hua  and
      Liu, Kang  and
      Che, Wanxiang  and
      He, Shizhu  and
      Rao, Gaoqi",
    booktitle = "Proceedings of the 20th Chinese National Conference on Computational Linguistics",
    month = aug,
    year = "2021",
    address = "Huhhot, China",
    publisher = "Chinese Information Processing Society of China",
    url = "https://aclanthology.org/2021.ccl-1.108/",
    pages = "1218--1227",
    language = "eng",
}

@inproceedings{deberta,
title={De{BERT}aV3: Improving De{BERT}a using {ELECTRA}-Style Pre-Training with Gradient-Disentangled Embedding Sharing},
author={Pengcheng He and Jianfeng Gao and Weizhu Chen},
booktitle={The Eleventh International Conference on Learning Representations },
year={2023},
url={https://openreview.net/forum?id=sE7-XhLxHA}
}

@inproceedings{gte-en-mlm,
    title = "{mGTE}: Generalized Long-Context Text Representation and Reranking Models for Multilingual Text Retrieval",
    author = "Zhang, Xin  and
      Zhang, Yanzhao  and
      Long, Dingkun  and
      Xie, Wen  and
      Dai, Ziqi  and
      Tang, Jialong  and
      Lin, Huan  and
      Yang, Baosong  and
      Xie, Pengjun  and
      Huang, Fei  and
      Zhang, Meishan  and
      Li, Wenjie  and
      Zhang, Min",
    editor = "Dernoncourt, Franck  and
      Preo{\c{t}}iuc-Pietro, Daniel  and
      Shimorina, Anastasia",
    booktitle = "Proceedings of the 2024 Conference on Empirical Methods in Natural Language Processing: Industry Track",
    month = nov,
    year = "2024",
    address = "Miami, Florida, US",
    publisher = "Association for Computational Linguistics",
    url = "https://aclanthology.org/2024.emnlp-industry.103/",
    doi = "10.18653/v1/2024.emnlp-industry.103",
    pages = "1393--1412",
}

@inproceedings{dayanik-etal-2022-improving,
    title = "Improving Neural Political Statement Classification with Class Hierarchical Information",
    author = "Dayanik, Erenay  and
      Blessing, Andre  and
      Blokker, Nico  and
      Haunss, Sebastian  and
      Kuhn, Jonas  and
      Lapesa, Gabriella  and
      Pado, Sebastian",
    editor = "Muresan, Smaranda  and
      Nakov, Preslav  and
      Villavicencio, Aline",
    booktitle = "Findings of the Association for Computational Linguistics: ACL 2022",
    month = may,
    year = "2022",
    address = "Dublin, Ireland",
    publisher = "Association for Computational Linguistics",
    url = "https://aclanthology.org/2022.findings-acl.186/",
    doi = "10.18653/v1/2022.findings-acl.186",
    pages = "2367--2382",
}

@inproceedings{3780338.3783035, author = {Wu, Xinyi and Wang, Yifei and Jegelka, Stefanie and Jadbabaie, Ali}, title = {On the emergence of position bias in transformers}, year = {2025}, booktitle = {Proceedings of the 42nd International Conference on Machine Learning}, articleno = {2697}, numpages = {26}, location = {Vancouver, Canada},
url = {https://openreview.net/forum?id=YufVk7I6Ii}}

@article{CARLSON_MONTGOMERY_2017, title={A Pairwise Comparison Framework for Fast, Flexible, and Reliable Human Coding of Political Texts}, volume={111}, DOI={10.1017/S0003055417000302}, number={4}, journal={American Political Science Review}, author={David Carlson and Jacob M. Montgomery}, year={2017}, pages={835–843}}

@inproceedings{flashattention,
  title={Flashattention: Fast and memory-efficient exact attention with {IO}-awareness},
  author={Dao, Tri and Fu, Dan and Ermon, Stefano and Rudra, Atri and R{\'e}, Christopher},
  booktitle={Advances in neural information processing systems},
  url = {https://proceedings.neurips.cc/paper_files/paper/2022/file/67d57c32e20fd0a7a302cb81d36e40d5-Paper-Conference.pdf},
  pages={16344--16359},
  year={2022}
}

@article{wagner2021party,
  title={The party politics of the {EU}’s relations with the {USA}: evidence from the {European Parliament}},
  author={Wagner, Wolfgang and Pelaez, Luis and Raunio, Tapio and van de Koppel, Maartje},
  journal={European security},
    url = {https://doi.org/10.1080/09662839.2021.1947803},
  volume={30},
  number={3},
  pages={418--438},
  year={2021},
  publisher={Taylor \& Francis}
}

@book{MarksSteenbergen2004,
  title={European integration and political conflict},
  author={Marks, Gary and Steenbergen, Marco R},
  year={2004},
  publisher={Cambridge University Press}
}

@article{jahn2011,
  title={Conceptualizing Left and Right in comparative politics: Towards a deductive approach},
  author={Jahn, Detlef},
  journal={Party Politics},
  volume={17},
  number={6},
  pages={745--765},
  url = {https://doi.org/10.1177/1354068810380091},
  year={2011},
  publisher={SAGE Publications Sage UK: London, England}
}

@article{albright2010,
  title={The multidimensional nature of party competition},
  author={Albright, Jeremy J},
  journal={Party Politics},
  volume={16},
  number={6},
  url = {https://journals.sagepub.com/doi/10.1177/1354068809345856},
  pages={699--719},
  year={2010},
  publisher={SAGE Publications Sage UK: London, England}
}

@article{jahn2023,
  title={The changing relevance and meaning of left and right in 34 party systems from 1945 to 2020},
  author={Jahn, Detlef},
  journal={Comparative European Politics},
  volume={21},
  url = {https://doi.org/10.1057/s41295-022-00305-5},
  number={3},
  pages={308--332},
  year={2023},
  publisher={Springer}
}

@book{bootstrap,
  title={An introduction to the bootstrap},
  author={Efron, Bradley and Tibshirani, Robert J},
  year={1994},
  publisher={Chapman and Hall/CRC}
}

@article{benoit2025llms,
  title={Using large language models to analyze political texts through natural language understanding},
  author={Benoit, Kenneth and De Marchi, Scott and Laver, Conor and Laver, Michael and Ma, Jinshuai},
  journal={American Journal of Political Science},
  year={2025},
  url = {https://onlinelibrary.wiley.com/doi/10.1111/ajps.70050},
  publisher={Wiley Online Library}
}

@article{ornstein2025llms,
  title={How to train your stochastic parrot: Large language models for political texts},
  author={Ornstein, Joseph T. and Blasingame, Elise N. and Truscott, Jake S.},
  journal={Political Science Research and Methods},
  volume={13},
  url = {https://doi.org/10.1017/psrm.2024.64},
  number={2},
  pages={264--281},
  year={2025},
  publisher={Cambridge University Press}
}

@article{lemens2025llms,
  title={Positioning political texts with large language models by asking and averaging},
  author={Le Mens, Ga{\"e}l and Gallego, Aina},
  journal={Political Analysis},
  volume={33},
  number={3},
  url = {https://www.cambridge.org/core/journals/political-analysis/article/positioning-political-texts-with-large-language-models-by-asking-and-averaging/BB17F58E7329B53BC4B5F4065C7E00FF},
  pages={274--282},
  year={2025},
  publisher={Cambridge University Press}
}

@misc{olmo2026olmo3,
      title={Olmo 3}, 
      author={Allyson Ettinger and Amanda Bertsch and Bailey Kuehl and David Graham and David Heineman and Dirk Groeneveld and Faeze Brahman and Finbarr Timbers and Hamish Ivison and Jacob Morrison and Jake Poznanski and Kyle Lo and Luca Soldaini and Matt Jordan and Mayee Chen and Michael Noukhovitch and Nathan Lambert and Pete Walsh and Pradeep Dasigi and Robert Berry and Saumya Malik and Saurabh Shah and Scott Geng and Shane Arora and Shashank Gupta and Taira Anderson and Teng Xiao and Tyler Murray and Tyler Romero and Victoria Graf and Akari Asai and Akshita Bhagia and Alexander Wettig and Alisa Liu and Aman Rangapur and Chloe Anastasiades and Costa Huang and Dustin Schwenk and Harsh Trivedi and Ian Magnusson and Jaron Lochner and Jiacheng Liu and Lester James V. Miranda and Maarten Sap and Malia Morgan and Michael Schmitz and Michal Guerquin and Michael Wilson and Regan Huff and Ronan Le Bras and Rui Xin and Rulin Shao and Sam Skjonsberg and Shannon Zejiang Shen and Shuyue Stella Li and Tucker Wilde and Valentina Pyatkin and Will Merrill and Yapei Chang and Yuling Gu and Zhiyuan Zeng and Ashish Sabharwal and Luke Zettlemoyer and Pang Wei Koh and Ali Farhadi and Noah A. Smith and Hannaneh Hajishirzi},
      year={2026},
      eprint={2512.13961},
      archivePrefix={arXiv},
      primaryClass={cs.CL},
      url={https://arxiv.org/abs/2512.13961}, 
}

@article{coder_reliability_2012,
  title={Coder reliability and misclassification in the human coding of party manifestos},
  author={Mikhaylov, Slava and Laver, Michael and Benoit, Kenneth R},
  journal={Political analysis},
  volume={20},
  number={1},
  url = {https://doi.org/10.1093/pan/mpr047},
  pages={78--91},
  year={2012},
  publisher={Cambridge University Press}
}

@article{Barbera_2015, 
    title={Birds of the Same Feather Tweet Together: Bayesian Ideal Point Estimation Using Twitter Data}, 
    volume={23}, 
    DOI={10.1093/pan/mpu011}, 
    number={1}, 
    journal={Political Analysis}, 
    author={Barberá, Pablo}, 
    year={2015}, pages={76–91}
}

@article{Diermeier_2012, 
    title={Language and Ideology in Congress}, 
    volume={42}, 
    DOI={10.1017/S0007123411000160}, 
    number={1}, 
    journal={British Journal of Political Science}, 
    author={Diermeier, Daniel and Godbout, Jean-François and Yu, Bei and Kaufmann, Stefan}, 
    year={2012}, 
    pages={31–55}
}

\appendix

\section{Experimental Setup}
\label{app:exp-setup}

\subsection{RILE and GAL-TAN Category Distribution}

\begin{figure}[hptb!]
    \centering
    \includegraphics[width=0.9\linewidth]{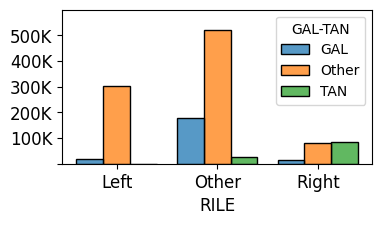}
    \caption{The distribution of the RILE and GAL-TAN category labels in the \textsc{x-time} dataset.}
    \label{fig:label-distribution}
\end{figure}

As shown in Fig.~\ref{fig:label-distribution}, the \textsc{x-time} dataset is largely skewed towards the category \textit{Other} for both RILE and GAL-TAN. All category combinations are present, except for \textit{Left+TAN}.

\subsection{Training Hyperparameters}
Following \citet{nikolaev-etal-2023}, all supervised models are trained for 5 epochs using the AdamW optimizer \citep{adamw} with early stopping. The learning rate is $10^{-5}$ for individual and joint multitask prediction, and $5 \cdot 10^{-5}$ for training the classification heads after joint contrastive tuning. The margin hyperparameter in the triplet loss function~(\ref{eq:triplet-loss}) is set to $1$ based on preliminary experiments. Following \citet{ceron-2022-dis-similarity}, we use Euclidean distance as the distance metric and employ a linear learning rate schedule with 100 warmup steps. The mini-batch size is 256 for label aggregation, except for contrastive tuning where it is set to 16, and 4 for chunk-level regression.

\subsection{LLM Prompts}

\paragraph{RILE} 
Question: What political position is expressed in this statement? \\
Statement: <sentence> \\
Option A: Right-wing \\
Option B: Left-wing \\
Option C: Neutral \\
Keep your response short (up to 10 words) by choosing exactly one option! \\
Correct option: \\

\paragraph{GAL-TAN} 
Question: What political position is expressed in this statement? \\
Statement: <sentence> \\
Option A: Green-Alternative-Liberal \\
Option B: Traditional-Authoritarian-Nationalist \\
Option C: Neutral \\
Keep your response short (up to 10 words) by choosing exactly one option! \\
Correct option: \\

\paragraph{Joint Prompting} 
Question: What political position is expressed in this statement? \\
Statement: <sentence> \\
Choose exactly one option from each of the two lists below. \\
List 1 (economic policy): \\
Option A: Right-wing \\
Option B: Left-wing \\
Option C: Neutral \\
List 2 (socio-cultural policy): \\
Option D: Green-Alternative-Liberal \\
Option E: Traditional-Authoritarian-Nationalist \\
Option F: Neutral \\
Keep your response short (up to 10 words)! \\
Correct options: \\

\subsection{Chunk Sizes}

Table~\ref{tab:sent-chunk-stats} gives an overview of the chunk sizes evaluated in the chunk size experiment on ModernBERT joint regression.

\begin{table}[!hptb]
\centering
\small
\begin{tabular}{@{}r|r|r|r|l|l@{}}
\toprule
\textbf{\begin{tabular}[c]{@{}r@{}}Max. \\ sent.\end{tabular}} & \multicolumn{1}{l|}{\textbf{\begin{tabular}[c]{@{}l@{}}Avg.\\ sent.\end{tabular}}} & \multicolumn{1}{l|}{\textbf{\begin{tabular}[c]{@{}l@{}}Max.\\ tokens\end{tabular}}} & \multicolumn{1}{l|}{\textbf{\begin{tabular}[c]{@{}l@{}}Avg.\\ tokens\end{tabular}}} & \multicolumn{1}{l|}{\textbf{\begin{tabular}[c]{@{}l@{}}Mini-\\batch\\ size\end{tabular}}} & \textbf{\begin{tabular}[c]{@{}l@{}}Train time / \\ epoch\\ (hh:mm)\footnotemark \end{tabular}} \\ \midrule
\textbf{1} & 1 & 735 & 23 & 32 & 02:40 \\
\textbf{5} & 5 & 1568 & 113 & 16 & 01:15 \\
\textbf{20} & 20 & 3351 & 451 & 16 & 00:40 \\
\textbf{50} & 49 & 4297 & 1117 & 8 & 00:40 \\
\textbf{100} & 97 & 5056 & 2206 & 4 & 00:40 \\
\textbf{200} & 189 & 8173 & 4281 & 2 & 00:50 \\
676 & 315 & \textbf{8192} & 7153 & 4 & 01:10 \\ \bottomrule
\end{tabular}
\caption{Chunk length statistics in the sentence-chunk smoothing experiment with ModernBERT. The values delimiting the chunk length are highlighted in bold.}
\label{tab:sent-chunk-stats}
\end{table}

\footnotetext{On a GPU server Nvidia RTX 6000 Ada, 48 GB.}

\subsection{Statistical Significance Testing}
Whenever comparing models with less than 15 percentage points of gap in performance, we run the \textit{bootstrap resampling test} to see if the difference is statistically significant \citep{bootstrap}. This test is non-parametric and thus applicable to any metric, including the rank correlation. Given two models $m_1$~and~$m_2$, the procedure is run as follows.

\begin{enumerate}
    \item The test set manifestos are sampled with replacement, the size of each sample the same as that of the original test set. The predictions of~$m_1$~and~$m_2$ for those manifestos as well as the ground truth scores are compiled accordingly.

    \item For each model and random seed, the rank correlation of the predictions with the ground truth is calculated. The resulting values are averaged over the random seeds, producing $\rho_1$ and $\rho_2$ to represent~$m_1$~and~$m_2$, respectively. Their difference $\rho_1 - \rho_2$ is the variable of interest.

    \item The operations in (1) and (2) are repeated $n=10,000$ times, creating a distribution of $\{\rho_1 - \rho_2\}$. 

    \item The $2,5$ and the $97,5$ percentile of that distribution are calculated, producing a $95\%$ confidence interval $[a,b]$. It is interpreted as follows:

    \begin{itemize}
        \item $0 \in [a,b]$ --- $m_1$~and~$m_2$ perform on a par
        
        \item $0 < a$ --- $m_1$ performs better than $m_2$
    
        \item $0 > b$ --- $m_1$ performs worse than $m_2$
    \end{itemize}
\end{enumerate}

\section{Extended Results}
\label{app:results}

\subsection{Label Aggregation}
Table~\ref{tab:results-label-aggr} presents a detailed evaluation of the label aggregation models,  additionally reporting accuracy, weighted F1-score and macro F1-score computed at the sentence classification level. The macro-F1 values are consistently lower than the accuracy and weighted F1 ($0.75 / 0.82$ vs $0.66/0.67$). Yet, among the classification-level metrics, macro-F1 is the most reliable predictor for the manifesto-level rank correlation $\rho$, which highlights the importance of evaluating all classes equally for a fair view of the model quality.

The SBERT and ModernBERT baselines have the encoder weights frozen and only the classification heads trained during fine-tuning. The baselines reveal that ModernBERT has a weaker starting point but catches up to SBERT when trained for individual or joint prediction. Notably, the SBERT baseline is more robust than the LLM baseline on both RILE and GAL-TAN.

The majority prediction baseline always outputs the label \textit{Other}, as it is the most prevalent in the training data for both RILE and GAL-TAN. For this method, the rank correlation cannot be calculated, since that requires dividing by the covariance which equals zero for a constant series (all of the manifesto-level estimates equal 1; cf. eq.~(\ref{eq:rile}, \ref{eq:galtan})).

\begin{table*}[b]
\centering
\small
\resizebox{\textwidth}{!}{%
\begin{tabular}{@{}llcccc|cccc@{}}
\toprule
 & \textbf{} & \multicolumn{4}{c|}{\textbf{RILE}} & \multicolumn{4}{c}{\textbf{GAL-TAN}} \\ \cmidrule(l){3-10} 
 &  & \textbf{acc} & \textbf{w. F1} & \textbf{m. F1} & $\bm{\rho}$ & \textbf{acc} & \textbf{w. F1} & \textbf{m. F1} & $\bm{\rho}$ \\ \midrule
- & Majority prediction & 0.63 & 0.49 & 0.26 & - & 0.79 & 0.69 & 0.29 & - \\ \midrule
\textbf{SBERT} & Baseline & 0.70 & 0.69 & 0.58 & 0.72 & 0.80 & 0.79 & 0.58 & 0.68 \\
 & Individual prediction & \textbf{0.75} & \textbf{0.75} & \textbf{0.66} & \textbf{0.88} & 0.82 & 0.82 & \textbf{0.67} & \textbf{0.83} \\
 & Joint multitask & \textbf{0.75} & \textbf{0.75} & \textbf{0.66} & \textbf{0.87} & 0.82 & \textbf{0.83} & \textbf{0.67} & \textbf{0.84} \\
 & Joint contrast. RLGT\textsubscript{whiten} & 0.73 & 0.74 & 0.65 & 0.83 & 0.80 & 0.81 & 0.64 & 0.77 \\ \midrule
\textbf{ModernBERT} & Baseline & 0.67 & 0.63 & 0.46 & 0.36 & 0.79 & 0.75 & 0.44 & 0.57 \\
 & Individual prediction & \textbf{0.75} & \textbf{0.75} & \textbf{0.66} & \textbf{0.87} & \textbf{0.83} & \textbf{0.83} & \textbf{0.67} & \textbf{0.83} \\
 & Joint multitask & 0.73 & 0.73 & 0.65 & \textbf{0.86} & 0.82 & 0.82 & 0.65 & \textbf{0.83} \\
 & Joint contrast. RLGT & 0.75 & 0.74 & 0.64 & 0.80 & 0.82 & 0.82 & 0.64 & 0.74 \\ \midrule
 \textbf{LLM (Olmo 3)} & Individual prediction & 0.49 & 0.50 & 0.38 & 0.67 & 0.50 & 0.56 & 0.35 & 0.53 \\
 & Joint prediction & 0.33 & 0.34 & 0.31 & 0.61 & 0.22 & 0.20 & 0.22 & 0.48 \\ \bottomrule
\end{tabular}%
}
\caption{The performance of label aggregation (primary overview). \textit{Joint contrast. RLGT} refers to joint contrastive tuning with RILE and GAL-TAN based triplets; \textit{whiten} indicates the whitening transformation. \textit{w. F1} and \textit{m. F1} are the weighted F1 and macro F1, respectively.}
\label{tab:results-label-aggr}
\end{table*}

\subsection{Joint Contrastive Training}
Table~\ref{tab:results-la-contr} allows a more detailed look into joint contrastive training. It achieves the rank correlation of at most $\rho=0.83$ and $\rho=0.77$ on RILE and GAL-TAN, respectively. The most robust way to select triplets for training is based on the RILE and GAL-TAN category labels. SBERT--RILE--GAL-TAN\textsubscript{whiten} and ModernBERT--RILE--GAL-TAN perform on a par in that setting. In particular, layering in the whitening transformation allows SBERT to get a small but statistically significant boost on RILE ($\rho=0.83$ vs $0.81$), whereas on GAL-TAN the difference in $\rho$ is insignificant. In contrast, the whitening transformation dramatically impairs the downstream classification quality of ModernBERT--RILE--GAL-TAN. 

Mining triplets based on party makes for inferior RILE and GAL-TAN classifiers. SBERT--Party, even with the aid of the whitening transformation, scores significantly lower even than the baseline SBERT where only the classification heads were trained. With slightly more success, ModernBERT--Party improves over the ModernBERT baseline on RILE and scores on a par on GAL-TAN. Again, whitening the embeddings causes a decrease in the ModernBERT scores.

\begin{table*}[!t]
\centering
\small
\resizebox{\textwidth}{!}{%
\begin{tabular}{@{}llcccc|cccc@{}}
\toprule
 &  & \multicolumn{4}{c|}{\textbf{RILE}} & \multicolumn{4}{c}{\textbf{GAL-TAN}} \\ \cmidrule(l){3-10} 
\textbf{} & \textbf{} & \textbf{acc} & \textbf{w. F1} & \textbf{m. F1} & $\bm{\rho}$ & \textbf{acc} & \textbf{w. F1} & \textbf{m. F1} & $\bm{\rho}$ \\ \midrule
SBERT & Baseline & 0.70 & 0.69 & 0.58 & 0.72 & 0.80 & 0.79 & 0.58 & 0.68 \\ \midrule
\multirow{4}{*}{\begin{tabular}[c]{@{}l@{}}SBERT \\ joint \\ contrastive\end{tabular}} & Party & 0.65 & 0.60 & 0.44 & 0.36 & 0.79 & 0.73 & 0.42 & 0.44 \\
 & Party\textsubscript{whiten} & 0.66 & 0.64 & 0.51 & 0.51 & 0.77 & 0.75 & 0.50 & 0.55 \\
 & RILE--GAL-TAN & 0.74 & \textbf{0.74} & \textbf{0.65} & 0.81 & \textbf{0.82} & \textbf{0.82} & \textbf{0.65} & \textbf{0.74} \\
 & RILE--GAL-TAN\textsubscript{whiten} & 0.73 & \textbf{0.74} & \textbf{0.65} & \textbf{0.83} & 0.80 & 0.81 & 0.64 & \textbf{0.77} \\ \toprule
ModernBERT & Baseline & 0.67 & 0.63 & 0.46 & 0.36 & 0.79 & 0.75 & 0.44 & 0.57 \\ \midrule
\multirow{4}{*}{\begin{tabular}[c]{@{}l@{}}ModernBERT\\ joint \\ contrastive\end{tabular}} & Party & 0.68 & 0.64 & 0.49 & 0.54 & 0.79 & 0.76 & 0.48 & 0.58 \\
 & Party\textsubscript{whiten} & 0.66 & 0.61 & 0.44 & 0.41 & 0.79 & 0.74 & 0.44 & 0.49 \\
 & RILE--GAL-TAN & \textbf{0.75} & \textbf{0.74} & 0.64 & \textbf{0.80} & \textbf{0.82} & \textbf{0.82} & 0.64 & \textbf{0.74} \\
 & RILE--GAL-TAN\textsubscript{whiten} & 0.63 & 0.49 & 0.26 & 0.17 & 0.79 & 0.69 & 0.29 & 0.08 \\ \bottomrule
\end{tabular}%
}
\caption{Performance of label aggregation with joint contrastive tuning. Apart from the baselines, the second column states the triplet mining strategy; \textit{whiten} is the whitening transformation.}
\label{tab:results-la-contr}
\end{table*}

\subsection{Chunk-Level Regression}

The chunk-level regression models are additionally assessed on the MSE value at the manifesto level. Its dynamics mostly correspond to those of the rank correlation score.

\begin{table*}[!hptb]
\centering
\begin{tabular}{@{}llcc|cl@{}}
\toprule
 &  & \multicolumn{2}{c|}{\textbf{RILE}} & \multicolumn{2}{c}{\textbf{GAL-TAN}} \\ \cmidrule(l){3-6}
 &  & \textbf{MSE} & $\bm{\rho}$ & \textbf{MSE} & $\bm{\rho}$ \\ \midrule
\textbf{BigBird} & Baseline & 0.018 & 0.61 & 0.015 & 0.58 \\
 & Individual prediction & 0.008 & \textbf{0.84} & 0.006 & \textbf{0.84} \\
 & Joint multitask & 0.009 & \textbf{0.84} & 0.007 & 0.82 \\ \midrule
\textbf{ModernBERT} & Baseline & \multicolumn{1}{l}{0.024} & \multicolumn{1}{l|}{0.45} & \multicolumn{1}{l}{0.015} & 0.33 \\
 & Individual prediction & \multicolumn{1}{l}{0.009} & \multicolumn{1}{l|}{0.79} & \multicolumn{1}{l}{0.007} & 0.78 \\
 & Joint multitask & \multicolumn{1}{l}{0.013} & \multicolumn{1}{l|}{\textbf{0.83}} & \multicolumn{1}{l}{0.008} & 0.79 \\ \bottomrule
\end{tabular}
\caption{Performance of chunk-level regression.}
\label{tab:results-regr}
\end{table*}

\end{document}